\documentclass[letterpaper, 10 pt, conference]{ieeeconf}  

\IEEEoverridecommandlockouts                              

\usepackage{amsmath}
\usepackage{booktabs}
\usepackage{array}
\usepackage{float}
\usepackage{graphicx}

\usepackage{xcolor}
\usepackage{hyperref}

\title{\LARGE \bf
Beyond Retrieval Relevance: Scene-Grounded Risk Entailment for Vision-Language Driving
}

\author{Jiaxin Liu$^{1,2}$, Ruilin Yu$^{1,3}$, Liang Peng$^{1}$, Jingkai Wang$^{1}$, Chengxiang Zhao$^{1,2}$, Zhenxin Zhu$^{2}$, Bing Wang$^{2}$,\\Guang Chen$^{2}$, Hangjun Ye$^{2}$, Hong Wang$^{1*}$ and Jun Li$^{1}$
\thanks{$^{1}$ School of Vehicle and Mobility, Tsinghua University}%
\thanks{$^{2}$ Xiaomi EV}%
\thanks{$^{3}$ Jilin University}%
}

\begin{document}

\maketitle
\thispagestyle{empty}
\pagestyle{empty}

\begin{abstract}

Retrieval-augmented generation (RAG) gives vision--language driving systems access to external safety knowledge, yet a retrieved risk rule may be relevant without applying to the current scene.
A vision--language model (VLM) receiving such knowledge must ground objects, bind entities across time, and verify relations before deciding how to act, leaving the support for risk conclusions implicit.
We address this relevance--applicability gap with a Driving-Risk Knowledge Graph (DRKG) and Semantic Web Rule Language (SWRL) reasoning stage before VLM decision-making.
Structured perception instantiates scene facts, from which SWRL rules derive events and directed risk relations when their antecedents are jointly satisfied.
Recognized events, bound risk relations, and semantic descriptions of activated rules form compact evidence that conditions the VLM and diffusion planner.
In matched comparisons on nuReasoning, our method improved the nuReasoning planning score (NPS) by 1.30 points and the non-at-fault collision score (NC) by 2.76 points over the relevance retrieval-based baseline.
These gains indicate that scene-applicable risk evidence improves safety-weighted planning relative to semantically retrieved risk knowledge.

\end{abstract}

\section{INTRODUCTION}

Safe driving requires an autonomous system to interpret current visual observations in light of traffic rules, commonsense risk knowledge, and prior experience \cite{liu2026enhancing}. Retrieval-augmented generation (RAG) supplements a model's parametric knowledge and immediate visual evidence with external information accessed at inference time \cite{lewis2020retrieval}. Driving applications have consequently used RAG for action explanation, decision prediction, and safety-oriented visual question answering \cite{yuan2024ragdriver,wang2025rad,ye2025safedriverag}, establishing semantic retrieval as an important route towards knowledge-augmented vision--language driving.

Semantic retrieval determines which knowledge is relevant to a scene, but relevance alone does not establish a scene-specific risk. A retrieved risk rule may be generally valid and match observed objects or behaviours, yet its conclusion requires the corresponding objects to be grounded, its entities and times to be bound, and its spatial, motion, and interaction relations to hold jointly. We call this distinction the \emph{relevance--applicability gap}: relevance identifies knowledge worth considering, whereas applicability requires the rule's encoded antecedents to be positively supported by the current scene.

In current RAG-based driving pipelines, retrieved knowledge is typically passed to a vision--language model (VLM), leaving this relevance-to-applicability pathway implicit within generative reasoning \cite{yuan2024ragdriver,wang2025rad,ye2025safedriverag}.
The VLM must associate visual objects with rule entities, bind temporal states, test multiple relations, and combine their antecedents.
It must then determine which entity poses risk to which other entity and decide how to act.
This burden grows with compositional risk complexity, while the support for an asserted risk is not exposed as an independently checkable result.
Graph-structured question answering, knowledge-graph-based behaviour prediction, and rule-filtered driving frameworks show that parts of this reasoning can be structured \cite{sima2024drivelm,hussien2025ragexplainable,wang2025hybriddriving}, but do not separate scene-supported risk entailment from downstream VLM decision-making.

\begin{figure}[htbp]
    \centering
    \includegraphics[trim=17bp 100bp 17bp 60bp, clip, width=0.98\linewidth]{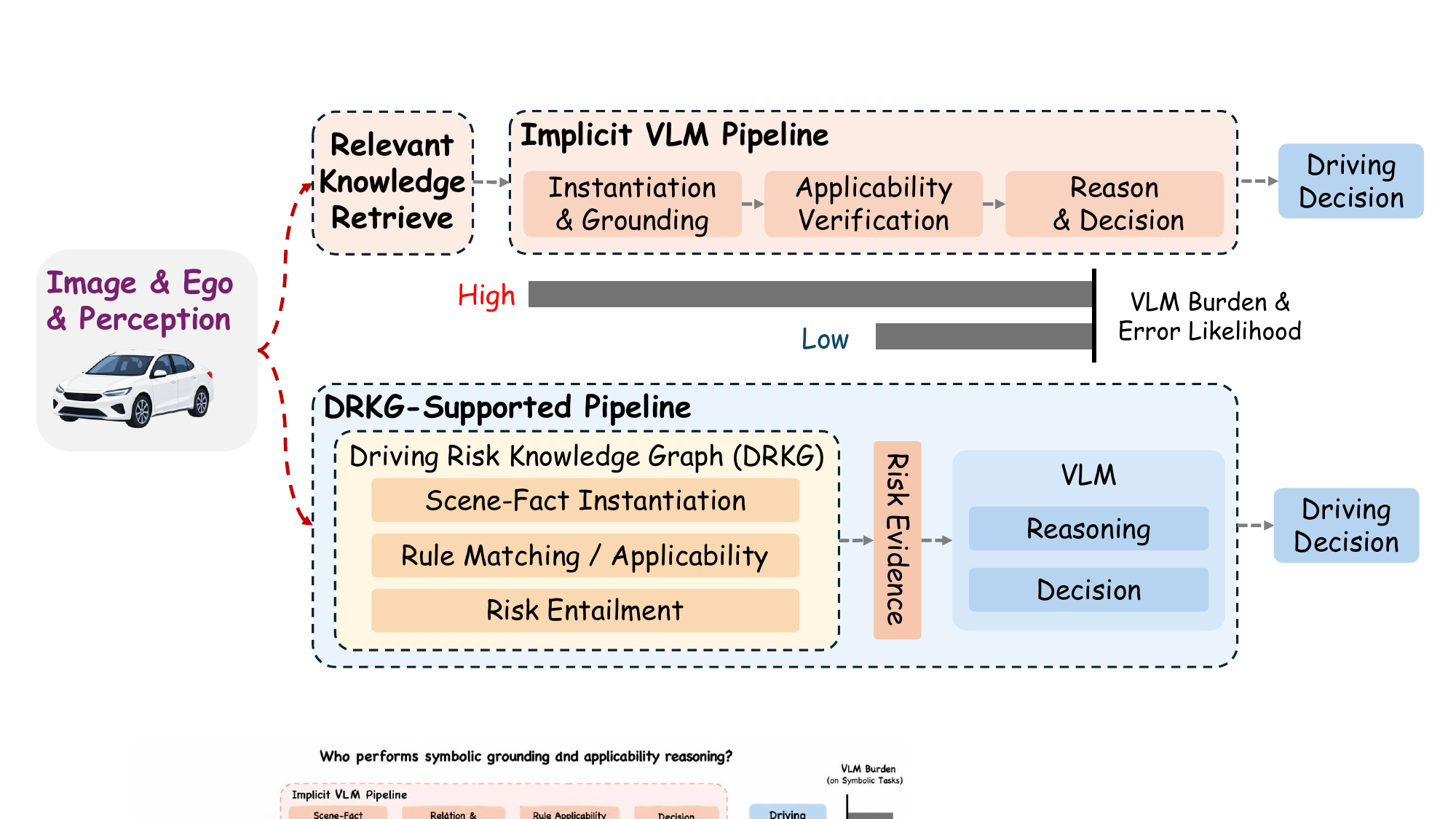}
    \caption{Typical RAG-based driving methods provide semantically relevant knowledge while leaving scene grounding and applicability assessment to the VLM, whereas our DRKG-supported framework explicitly derives scene-grounded risk evidence before VLM reasoning and decision making.}
    \label{fig:intro}
\end{figure}

We replace this retrieval-conditioned relevance-to-applicability pathway with scene-fact instantiation and symbolic risk reasoning over a Driving-Risk Knowledge Graph (DRKG), as shown in Fig.~\ref{fig:intro}. 
Existing perception outputs initialize scene entities, measured attributes, and simple spatial and kinematic relations.
Semantic Web Rule Language (SWRL) reasoning then considers all encoded rules and composes these direct facts into scene events and more complex relations.
We collect the recognized events together with each inferred risk relation and the risk rule that identifies its type as \emph{scene-grounded risk evidence}.
The VLM integrates a compact semantic serialization of this evidence with camera observations and ego context for decision-making and diffusion-based trajectory planning.

We evaluate this distinction through controlled comparisons on nuReasoning \cite{huang2026nureasoning}, varying only the supplied risk information.
Across node- and text-based retrieval strategies, expanding the candidate set recovers more activated risk rules but also introduces more scene-inapplicable rules.
The added matches do not reliably improve planning, whereas scene-grounded risk evidence yields better safety-weighted planning, particularly in non-at-fault collision avoidance.
Comparisons of evidence representations further show that scene-bound results and activated rule descriptions together yield a higher planning score than either alone.

Our contributions are:
\begin{itemize}
    \setlength{\itemsep}{0pt}
    \setlength{\parskip}{0pt}
    \setlength{\parsep}{0pt}
    \item We formulate the relevance-applicability gap in knowledge-augmented vision-language driving, showing that semantically relevant risk knowledge cannot be treated as scene-supported evidence without explicit grounding, binding, relational verification, and antecedent composition.
    \item We construct scene-grounded risk evidence by instantiating structured perception as ontology-grounded DRKG facts and applying closure-based SWRL reasoning.
    The evidence combines recognized events, directed risk relations between bound risk sources and targets, and semantic descriptions of activated rules for downstream VLM reasoning and planning.
    \item We demonstrate the downstream value of scene applicability over semantic relevance. Controlled planning comparisons show that scene-grounded risk evidence improves the safety-weighted planning over relevance-retrieved risk knowledge.
\end{itemize}
\section{RELATED WORK}

\subsection{Knowledge Access for Vision--Language Driving}

Retrieval-augmented driving has shown that external knowledge can support explanation, decision-making and planning.
RAG-Driver retrieves expert demonstrations for action explanation and prediction, RAD retrieves driving experience for meta-action selection, and SafeDriveRAG retrieves safety-knowledge subgraphs for visual question answering \cite{yuan2024ragdriver,wang2025rad,ye2025safedriverag}.
KnowVal uses scene-derived keywords to retrieve knowledge-graph entities, expands their neighbouring nodes and filters the resulting driving clauses for value-guided trajectory assessment \cite{xia2026knowval}.
DriveReg retrieves traffic-regulation paragraphs by scene-conditioned text similarity, refines the match at the sentence level and uses an LLM reasoning agent to assess rule applicability and action compliance \cite{cai2026drivereg}.
These pipelines access related knowledge through distinct mechanisms, but none exposes a risk relation between bound scene entities that has been formally entailed from scene facts.

\subsection{Structured Knowledge and Formal Risk Reasoning}

Research on structured and symbolic driving has shown that scene entities, temporal states, relations and formal conditions can be represented and evaluated outside a VLM.
DriveLM represents dependencies among perception, prediction and planning through graph-structured question answering, while Hybrid-Driving combines a scenario-evolution knowledge graph with rule-filtered action selection \cite{sima2024drivelm,wang2025hybriddriving}.
Formal systems also make condition checking explicit.
SGSM++ synthesizes runtime monitors from temporal safety properties over scene graphs, while Toledo \emph{et al.} monitor such properties over spatial relations extracted by a VLM \cite{woodlief2025sgsm,toledo2025monitoring}.
Their resources and interfaces nevertheless remain task-specific.
OD-RASE focuses on accident-causing road structures, whereas formal monitors evaluate predefined safety properties \cite{shimomura2025odrase}.
Together, these studies provide the representations and reasoning mechanisms required for external formal risk reasoning.
However, an executable interface is still needed to determine applicability across heterogeneous risks and return recognized events together with object-bound, type-identified risk relations.

\section{METHOD}

\begin{figure*}[thbp]
    \centering
    \includegraphics[trim=1bp 87bp 12bp 37bp, clip, width=0.98\linewidth]{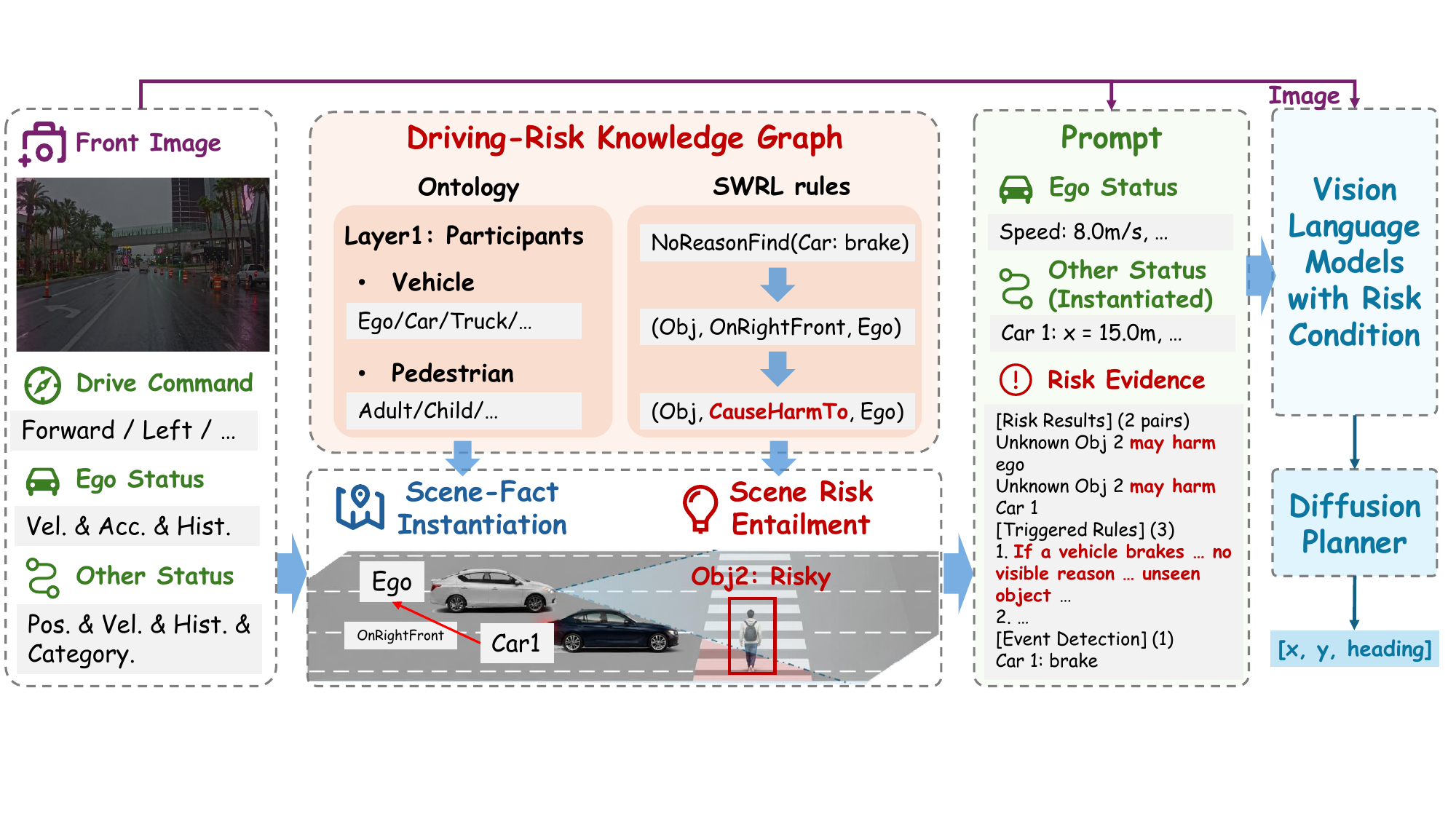}
    \caption{Overview of the scene-grounded risk-evidence vision language-driving pipeline.
    Structured perception results and ego state instantiate scene facts using the ontology of the driving-risk knowledge graph (DRKG).
    Its Semantic Web Rule Language (SWRL) rules derive scene events and risk relations between scene entities.
    Recognized events, risk relations and associated rule descriptions are combined with ego and scene context in a prompt, while camera observations provide the visual input to the vision--language model (VLM).
    The VLM decision representation conditions a diffusion planner to generate the ego trajectory.}
    \label{fig:pipeline}
    \vspace{-0.3cm}
\end{figure*}

\subsection{Problem Formulation}
For a driving scene $s$, let $\mathcal{O}_s$ denote camera observations, $\mathcal{P}_s$ structured perception results, and $\mathbf{x}_s$ the ego state.
Let $\mathcal{G}$ denote the fixed Driving-Risk Knowledge Graph (DRKG).
The task is to generate a planned ego trajectory $\boldsymbol{\tau}_s$.

To address the relevance--applicability gap, we derive scene-grounded risk evidence $\mathcal{Z}_s$ from structured perception and ego state using $\mathcal{G}$.
The resulting evidence conditions VLM decision-making and diffusion planning of $\boldsymbol{\tau}_s$ alongside camera observations and structured scene inputs.
We express these dependencies as
\begin{equation}
    \mathcal{Z}_s = H\!\left(\mathcal{P}_s,\mathbf{x}_s ; \mathcal{G}\right),
    \qquad
    \boldsymbol{\tau}_s = \Pi\!\left(\mathcal{O}_s,\mathbf{x}_s,\mathcal{P}_s,\mathcal{Z}_s\right),
\end{equation}

Here, $H$ denotes \emph{scene-applicability reasoning}, which instantiates scene facts, applies ontology and SWRL entailment, and extracts recognized events and inferred risk relations paired with their deriving rules.
The relations bind risk sources and targets in the scene, while the paired rules identify the risk types.
$\Pi$ denotes \emph{risk-evidence-conditioned decision and planning}, integrating VLM decisions with diffusion-based trajectory generation.
Within $\Pi$, the VLM integrates $\mathcal{Z}_s$ with visual and ego context, and its decision representation guides the diffusion planner.
The whole sequence, shown in Fig.~\ref{fig:pipeline}, places scene-applicability verification in $H$ before VLM decision-making, avoiding the misleading of relevant yet inapplicable rules.

\subsection{Scene-applicability Reasoning}
\subsubsection{Driving-Risk Knowledge Graph}
We represent the driving-risk knowledge base as
\begin{equation}
    \mathcal{G}=\left(\mathcal{T},\mathcal{R}_{\mathcal{G}}\right),
\end{equation}
where $\mathcal{T}$ is an OWL~2 ontology and $\mathcal{R}_{\mathcal{G}}$ is a set of positive SWRL rules defined over $\mathcal{T}$ \cite{hitzler2012owl2primer,horrocks2004swrl}.
The ontology fixes what can be represented, while the rules specify how additional assertions can be derived from scene facts.

Building on ontology-based traffic-scene modelling \cite{buechel2017ontology,huang2019ontology}, $\mathcal{T}$ is organized into five connected modules.
These modules represent (i) static scene entities and road topology, including the ego vehicle, traffic participants and scene regions, (ii) temporal identity and frame order, (iii) temporal and motion events and activities, (iv) unexpected events and their causes, and (v) risk phenomena.
Spatial, kinematic, occlusion, topological, behavioural and interaction relations connect entities within and across the modules.
Data properties record measurable states such as position and velocity.
The risk vocabulary defines one binary predicate, $\lambda_{\mathrm{risk}}$.
For scene entities $e_i$ and $e_j$, $\lambda_{\mathrm{risk}}(e_i,e_j)$ states that the head entity $e_i$ poses a risk to the tail entity $e_j$.

The DRKG covers simple kinematic risks, cut-in and crossing interactions, road-structure and right-of-way risks, occlusion-related risks, and complex compositional risks.
Simple risks depend on a small number of motion conditions, whereas compositional risks combine multiple entities, temporal states and relations across ontology modules.
We denote by $\mathcal{R}_{\mathrm{risk}}\subseteq\mathcal{R}_{\mathcal G}$ the risk rules whose conclusions instantiate $\lambda_{\mathrm{risk}}$.
These rules express different risk forms while sharing the same conclusion predicate.


Each SWRL rule $r\in\mathcal{R}_{\mathcal{G}}$ comprises an antecedent and a conclusion, each written as a conjunction of positive atomic propositions.
The antecedent specifies the entity, temporal, attribute and relational conditions that must be satisfied before the conclusion can be derived.
Intermediate rules may conclude a scene event or a complex relation.
A risk rule $r\in\mathcal{R}_{\mathrm{risk}}$ concludes $\lambda_{\mathrm{risk}}(e_i,e_j)$ for two traffic participant entities.
Deriving this conclusion establishes that every antecedent of $r$ is satisfied in the current scene.
The head and tail of $\lambda_{\mathrm{risk}}(e_i,e_j)$ identify the risk source and target, while the identity of $r$ specifies the risk type.
The construction procedure and category-level coverage are detailed in the Supplementary Material.

\begin{figure*}[thbp]
    \centering
    \includegraphics[trim=9bp 75bp 12bp 42bp, clip, width=0.98\linewidth]{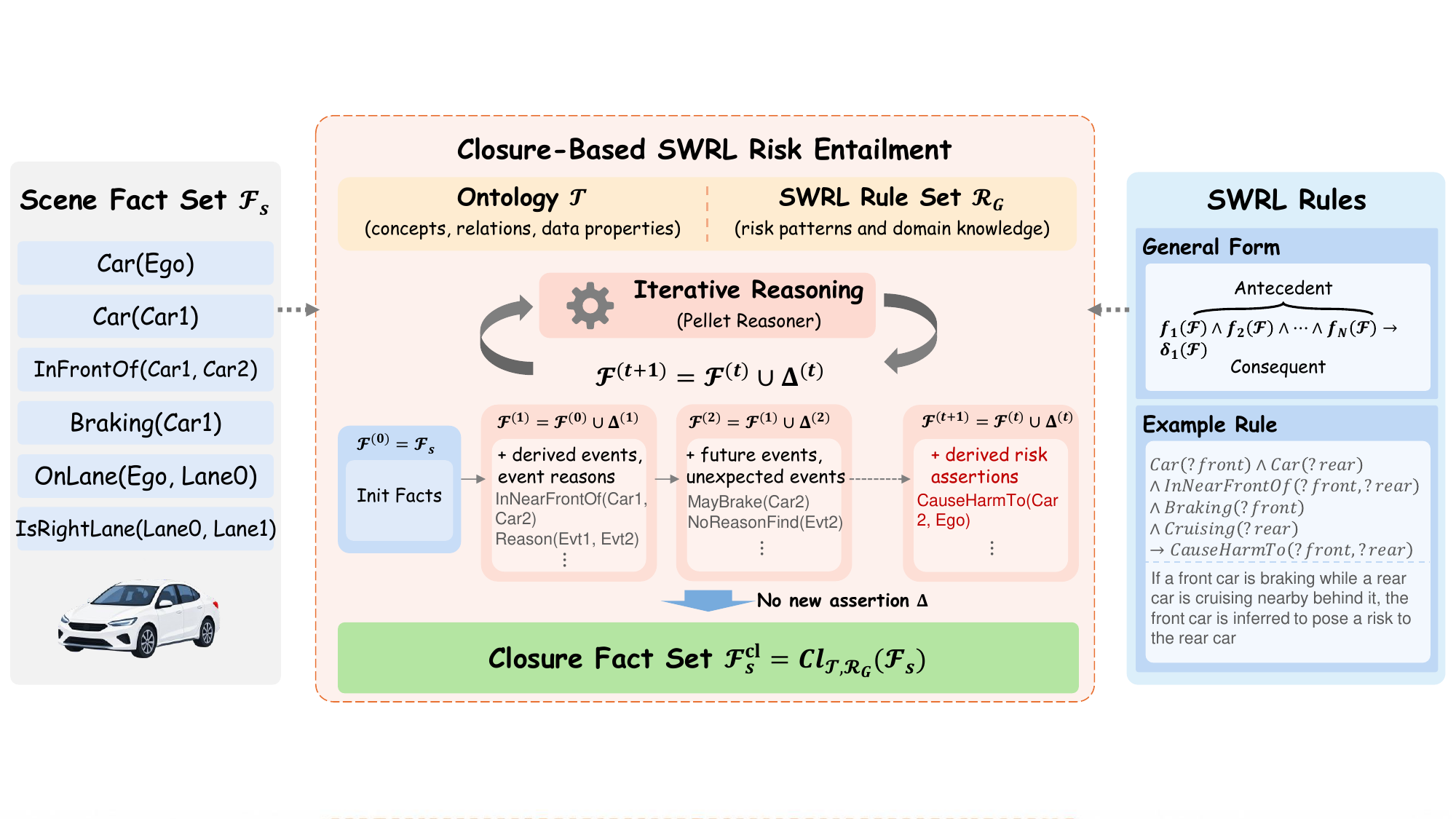}
    \caption{Closure-based SWRL risk entailment from scene facts.
    The initial fact set $\mathcal{F}_s$ is expanded by applying the ontology $\mathcal{T}$ and SWRL rules $\mathcal{R}_{\mathcal G}$ until no new assertion is derived.
    Intermediate events and relations lead to risk assertions in the closure $\mathcal{F}_s^{\mathrm{cl}}$.
    The right panel shows the general rule form and an illustrative vehicle-interaction rule.}
    \label{fig:swrl}
    \vspace{-0.3cm}
    
\end{figure*}

\subsubsection{Scene-Fact Instantiation}
Scene-fact instantiation converts structured perception into scene facts expressed with the ontology vocabulary.
Given perception results $\mathcal{P}_s$ and ego state $\mathbf{x}_s$, a scene adapter $g$ initializes the directly obtainable fact set
\begin{equation}
    \mathcal{F}_s
    =g\!\left(\mathcal{P}_s,\mathbf{x}_s;\mathcal{T}\right)
\end{equation}
The set $\mathcal{F}_s$ contains entity and class assertions, measured attributes, and simple relations supported directly by the perception output.
The rule set $\mathcal{R}_{\mathcal G}$ is not applied at this stage.

The adapter initializes the ego vehicle and each detected or tracked traffic participant as a scene-local individual.
It also initializes the road and lane entities represented in the structured inputs.
Stable numeric identifiers preserve participant identity across frames, while time-indexed attributes record quantities such as position, velocity, acceleration and heading.

Some simple spatial and kinematic relations are asserted in $\mathcal{F}_s$.
They are read from structured fields or computed deterministically from measured states and their temporal changes.
These relations include relative positions, basic motion states and manuevers without assigning a hazard interpretation.
Instances of $\lambda_{\mathrm{risk}}(e_i,e_j)$ and other complex relations are derived in the subsequent SWRL stage.

\subsubsection{SWRL-Based Scene Risk Entailment}

Scene risk entailment uses the SWRL rules in $\mathcal{R}_{\mathcal G}$ to derive scene events and complex relations from $\mathcal{F}_s$, ultimately producing instances of $\lambda_{\mathrm{risk}}(e_i,e_j)$ \cite{horrocks2004swrl}.
Mathematically, $\operatorname{Cl}_{\mathcal{T},\mathcal{R}_{\mathcal G}}$ denotes the closure operator induced by the ontology $\mathcal{T}$ and rule set $\mathcal{R}_{\mathcal G}$.
Starting from $\mathcal{F}_s$, it repeatedly applies the ontology and SWRL rules and adds each newly derived assertion to the current fact set until no further assertion is produced.
Fig.~\ref{fig:swrl} illustrates this progression from initial facts through intermediate assertions to a scene-specific risk relation and fianl SWRL closure fact set, and the process can be expressed as
\begin{equation}
    \mathcal{F}_s^{\mathrm{cl}}
    =\operatorname{Cl}_{\mathcal{T},\mathcal{R}_{\mathcal G}}
    \!\left(\mathcal{F}_s\right)
\end{equation}
$\mathcal{F}_s^{\mathrm{cl}}$ therefore contains both the initial facts in $\mathcal{F}_s$ and all facts derived from them.
The vocabulary remains fixed by $\mathcal{T}$.
We implement this entailment process using the Pellet reasoner \cite{sirin2007pellet}.

\subsubsection{Scene-Grounded Risk Evidence}
After SWRL entailment, we extract scene-grounded risk evidence $\mathcal{Z}_s$ for downstream decision-making.
It contains recognized scene events and inferred risk relations paired with the rules that derive them.
The event set $\mathcal{E}_s$ comprises temporal, motion and unexpected events recognized by the third and fourth ontology modules.
We write
\begin{equation}
    \mathcal{Z}_s
    =\left\langle\mathcal{E}_s,
    \left\{\left(\lambda_{\mathrm{risk}}(e_i,e_j),r\right)\right\}
    \right\rangle
    \label{eq:risk_evidence}
\end{equation}
The second component includes a pair for each inferred relation and every rule that derives it in scene $s$.
The relation identifies the risk source and target, while the rule identifies the risk type.

For VLM input, we serialize the recognized events and inferred risk relations, with predefined semantic descriptions mapping to replace the formal SWRL syntax of the associated risk rules with human-readable text.
This representation avoids requiring the VLM to parse complex conjunctions of ontology atoms or repeat low-level kinematic facts already available from perception.

\subsection{Risk-Evidence-Conditioned Decision and Planning}
\subsubsection{Risk-Evidence-Conditioned VLM Reasoning}
The visual input $\mathcal{O}_s$ comprises the current camera frame and four historical frames sampled at 2 Hz.
The VLM also receives ego and scene context $\mathbf{c}_s$ and scene-grounded risk evidence $\mathcal{Z}_s$.
The two textual sources are concatenated using a fixed prompt template $T$.
The resulting textual prompt is
\begin{equation}
    \mathbf{p}_s
    =T\!\left(\mathbf{c}_s,
   \mathcal{Z}_s\right)
\end{equation}

Conditioned on $\mathcal{O}_s$ and $\mathbf{p}_s$, the VLM produces a four-level output covering the scene, events, risks and driving decision.
We express this hierarchical output as
\begin{equation}
    \mathbf{y}_s
    =V_{\theta}\!\left(\mathcal{O}_s,\mathbf{p}_s\right)
    =\left(
    \mathbf{y}_s^{\mathrm{scene}},
    \mathbf{y}_s^{\mathrm{event}},
    \mathbf{y}_s^{\mathrm{risk}},
    \mathbf{y}_s^{\mathrm{decision}}
    \right)
\end{equation}
where $V_{\theta}$ denotes the VLM.
The decision level terminates with a longitudinal meta-action token, a lateral meta-action token and a plan token.
Let $\mathbf{h}_s^{\mathrm{lon}}$, $\mathbf{h}_s^{\mathrm{lat}}$ and $\mathbf{h}_s^{\mathrm{plan}}$ denote their final-layer hidden states.
Their ordered tuple forms the decision representation
\begin{equation}
    \mathbf{d}_s
    =\left(
    \mathbf{h}_s^{\mathrm{lon}},
    \mathbf{h}_s^{\mathrm{lat}},
    \mathbf{h}_s^{\mathrm{plan}}
    \right)
\end{equation}
which is passed to the diffusion planner as its VLM-derived conditioning signal.


\subsubsection{Conditioned Diffusion Planning}
Because our VLM primarily reasons about risk and high-level driving intent, we provide the diffusion planner with complementary geometric and dynamic context.
We adapt the ReCogDrive diffusion planner \cite{li2026recogdrive} to condition trajectory generation on this context and on $\mathbf{d}_s$, which carries the VLM's longitudinal, lateral and planning signals.
We then extract bird's-eye-view (BEV) scene features using the WorldEngine encoder \cite{li2026worldengine}, and encode surrounding agents and ego motion separately.
The resulting conditions are
\begin{equation}
    \begin{aligned}
    \mathbf{b}_s&=W_{\mathrm{BEV}}(\mathcal{O}_s),\\
    \mathbf{a}_s&=E_{\mathrm{agent}}(\mathcal{P}_s),\\
    \mathbf{u}_s&=E_{\mathrm{ego}}(\mathbf{x}_s,\mathbf{x}_s^{\mathrm{hist}}).
    \end{aligned}
    \label{eq:planner_conditions}
\end{equation}
where $\mathbf{x}_s^{\mathrm{hist}}$ denotes the historical ego states, $\mathbf{b}_s$ the BEV features, $\mathbf{a}_s$ the surrounding-agent features and $\mathbf{u}_s$ the ego-motion features.

Starting from a noisy trajectory $\boldsymbol{\tau}_s^{(K)}$, the planner performs conditional denoising at each step $k$,
\begin{equation}
    \boldsymbol{\tau}_s^{(k-1)}
    =D_{\phi}\!\left(\boldsymbol{\tau}_s^{(k)},k
    \mid\mathbf{d}_s,\mathbf{b}_s,\mathbf{a}_s,\mathbf{u}_s\right),
    \qquad
    \boldsymbol{\tau}_s=\boldsymbol{\tau}_s^{(0)}.
    \label{eq:conditional_planning}
\end{equation}
Here $D_{\phi}$ denotes a denoising update, and $\boldsymbol{\tau}_s$ is the planned ego trajectory.
The VLM hidden states guide trajectory refinement, BEV features enter through trajectory cross-attention, and the agent and ego encodings provide additional conditions.

\section{Experiments}

To examine the relevance--applicability gap in driving, we ask whether retrieved risk rules support scene-specific conclusions and whether this distinction affects planning.
Within the DRKG, SWRL activations provide the formal reference for measuring recall and applicability of risk-concluding rules and the rules along their derivation chains.
On nuReasoning, controlled planning comparisons vary the supplied risk information, and an evidence ablation separates the contributions of inferred results and activated rule descriptions.
A qualitative occlusion case illustrates the resulting risk reasoning and driving response.

\subsection{Dataset and Evaluation Metrics}
\subsubsection{Evaluation Dataset}
We use nuReasoning dataset because its long-tail driving scenes involve spatial relations and agent interactions, and its planning benchmark allows us to evaluate the downstream effect of risk evidence \cite{huang2026nureasoning}.
We split the currently available training portion into 1,943 training, 259 validation and 387 test scenes, approximating a 75:10:15 split.
Our local test split differs from the official benchmark test set (which is unavailable now), so published scores on that set provide context rather than a direct comparison.

\subsubsection{Metrics}
Within the DRKG, we use the SWRL rules that contribute to each inferred scene risk as the retrieval reference.
We evaluate retrieval of risk rules that conclude $\lambda_{\mathrm{risk}}$ and of all rules activated along the inference chains leading to those conclusions.
Specifically, we assess retrieval quality using recall and applicability.

\begin{itemize}
    \item \textbf{Recall: Are risk-supporting rules retrieved?}
    We average the fraction of reference rules retrieved over scenes with an inferred risk.
    \item \textbf{Applicability: Do retrieved rules support an inferred risk?}
    We average the fraction of retrieved rules in the reference over scenes with nonempty retrieval, including those without an inferred risk.
\end{itemize}

For planning, we adopt the nuReasoning Planning Score (NPS), which combines five normalized component scores as follows \cite{huang2026nureasoning}.
\begin{equation}
\mathrm{NPS}
=s_{\mathrm{NC}}s_{\mathrm{DA}}
\left(0.3s_{\mathrm{EP}}+0.2s_{\mathrm{CF}}+0.5s_{\mathrm{HL}}\right),
\label{eq:nps}
\end{equation}
where NC and DA denote non-at-fault collision and driveable-area compliance, respectively, and act as multiplicative safety gates.
EP, CF, and HL denote ego progress, comfort, and human-likeness, respectively.
Following nuReasoning, we also report average displacement error (ADE) over a five-second planning horizon.

\subsection{Implementation Details}
\textbf{Vision-Language Model.}
We initialize the VLM from Qwen3-VL-8B \cite{bai2025qwen3vl}.
We first conduct driving-domain pretraining on question-answer pairs from LingoQA and CODA-LM \cite{marcu2024lingoqa,chen2025automated}.
This stage provides supervision for interpreting consecutive driving frames and acquiring driving-risk knowledge.
We then use GPT-5.4 to align training samples from the nuReasoning training split and PotentialRiskQA with the four-level output format defined above \cite{huang2026nureasoning,liu2025potentialrisk}.
Supervised fine-tuning on the aligned samples trains the VLM to generate risk-focused reasoning and the longitudinal meta-action, lateral meta-action and plan tokens whose hidden states condition the diffusion planner.

\textbf{Diffusion Planner.}
We train the planner with a proximity-weighted objective that combines trajectory accuracy, command consistency, comfort and collision avoidance.
For $N$ training clips, the objective is
\begin{equation}
\begin{aligned}
\mathcal{L}_{\mathrm{DP}}
=\frac{1}{N}\sum_{s=1}^{N}w_s\bigl(&
\omega_{\mathrm{traj}}\mathcal{L}_{\mathrm{traj}}^{(s)}
+\omega_{\mathrm{cmd}}\mathcal{L}_{\mathrm{cmd}}^{(s)}\\
&+\omega_{\mathrm{comf}}\mathcal{L}_{\mathrm{comf}}^{(s)}
+\omega_{\mathrm{coll}}\mathcal{L}_{\mathrm{coll}}^{(s)}\bigr),
\end{aligned}
\label{eq:planner_loss}
\end{equation}
where $w_s$ is the agent-proximity weight for scene $s$ and the weights $\omega$ terms balance the four losses.

The trajectory loss supervises predicted future waypoints and endpoint states against the reference trajectory.
The command-consistency loss constrains terminal lateral displacement and heading to agree with the VLM's high-level driving command.
The comfort loss penalizes dynamically undesirable trajectories, while the collision loss penalizes collision between the ego vehicle and surrounding agents.
We assign larger $w_s$ to scenes with nearby traffic participants to emphasize safety-critical interactions during training.

Planner training proceeds in two stages.
We first train the VLM-conditioned diffusion backbone to map the decision representation $\mathbf{d}_s$ to future ego trajectories.
Then, we add the BEV, surrounding-agent and ego-history conditions in Eq.~\eqref{eq:planner_conditions} and further fine-tune the planner.
The second stage thus refines an established VLM-to-trajectory mapping with geometric and dynamic scene information.


\begin{table}[ht]
\caption{Rule recall and applicability (Appl.) against SWRL-derived scene references (\%).}
\label{tab:rule_retrieval}
\centering
\footnotesize
\setlength{\tabcolsep}{3pt}
\begin{tabular}{lcccc}
\toprule
Method & \multicolumn{2}{c}{Risk rules} & \multicolumn{2}{c}{Risk chains} \\
\cmidrule(lr){2-3}\cmidrule(lr){4-5}
 & Recall $\uparrow$ & Appl. $\uparrow$ & Recall $\uparrow$ & Appl. $\uparrow$ \\
\midrule
KnowVal top-5 & 18.32 & 11.11 & 6.52 & 1.09 \\
DriveReg top-5 & 29.30 & 12.58 & 27.05 & 1.55 \\
KnowVal top-16 & 65.57 & 8.05 & 50.73 & 1.11 \\
DriveReg top-16 & 33.88 & 4.04 & 29.51 & 0.58 \\
\midrule
SWRL-activated & \textbf{100.00} & \textbf{100.00} & \textbf{100.00} & \textbf{100.00} \\
\bottomrule
\end{tabular}
\end{table}

\begin{table*}[t]
\caption{Planning results with different risk-information sources.}
\label{tab:planning_retrieval}
\centering
\footnotesize
\setlength{\tabcolsep}{6pt}
\begin{tabular}{lccccccc}
\toprule
Risk information & NC $\uparrow$ & DA $\uparrow$ & EP $\uparrow$ & CF $\uparrow$ & HL $\uparrow$ & NPS $\uparrow$ & ADE $\downarrow$ \\
\midrule
KnowVal top-5 & 86.88 & 92.91 & 89.94 & 89.76 & 61.00 & 62.01 & 1.675 \\
DriveReg top-5 & 87.27 & \textbf{93.70} & 89.91 & 89.76 & 60.76 & 62.53 & 1.676 \\
KnowVal top-16 & 87.53 & 93.18 & \textbf{89.95} & \textbf{90.29} & \textbf{61.59} & 62.88 & \textbf{1.644} \\
DriveReg top-16 & 87.14 & 93.18 & 89.59 & 89.24 & 61.11 & 62.49 & 1.681 \\
Scene-grounded risk evidence (\textbf{ours}) & \textbf{90.29} & 92.91 & 89.79 & \textbf{90.29} & 60.17 & \textbf{64.18} & 1.728 \\
\bottomrule
\end{tabular}
\end{table*}

\begin{table*}[t]
\caption{Planning ablation of scene-grounded risk-evidence representation.}
\label{tab:rule_text_ablation}
\centering
\footnotesize
\setlength{\tabcolsep}{6pt}
\begin{tabular}{lccccccc}
\toprule
Risk-evidence input & NC $\uparrow$ & DA $\uparrow$ & EP $\uparrow$ & CF $\uparrow$ & HL $\uparrow$ & NPS $\uparrow$ & ADE $\downarrow$ \\
\midrule
None & 86.22 & \textbf{92.91} & \textbf{90.19} & 89.24 & 60.78 & 61.55 & \textbf{1.700} \\
Inferred results & 89.24 & 92.65 & 89.62 & \textbf{90.55} & \textbf{61.11} & 63.22 & 1.707 \\
Activated rule descriptions & 89.11 & 92.39 & 89.63 & 90.29 & 60.65 & 62.68 & 1.718 \\
Inferred results and rule descriptions (\textbf{ours}) & \textbf{90.29} & \textbf{92.91} & 89.79 & 90.29 & 60.17 & \textbf{64.18} & 1.728 \\
\bottomrule
\end{tabular}
\vspace{-0.3cm}
\end{table*}

\textbf{Baselines.}
The KnowVal-style baseline adapts keyword-based node matching and graph expansion \cite{xia2026knowval}.
We index each DRKG rule as a node using keywords and link nodes with similar keywords.
VLM-extracted scene keywords retrieve seed nodes and their neighbours.
The DriveReg-style baseline adapts paragraph-then-sentence text retrieval \cite{cai2026drivereg}.
We group semantically similar rules into paragraphs, match them against a VLM-generated scene summary, and then retrieve individual rules within the selected paragraphs.
Both baselines search the same DRKG rule base used for entailment at top-5 and top-16.
Planning comparisons hold camera observations, ego and scene context, the VLM, and the diffusion planner fixed while varying only the supplied risk information.

\subsection{Rule Retrieval and Applicability}

Broader semantic retrieval improved recall of rules concluding $\lambda_{\mathrm{risk}}$ but reduced their scene applicability (Table~\ref{tab:rule_retrieval}).
KnowVal recall rose from 18.32\% at top-5 to 65.57\% at top-16, while applicability fell from 11.11\% to 8.05\%.
DriveReg showed the same pattern, with recall rising from 29.30\% to 33.88\% and applicability falling from 12.58\% to 4.04\%.
Wider retrieval therefore covered more activated risk rules while returning a larger share of rules unsupported by the current scene.

Chain recall remained below risk-rule recall in every baseline setting, including 50.73\% versus 65.57\% for KnowVal top-16.
Chain applicability was only 0.58\%--1.55\%, compared with 4.04\%--12.58\% for rules concluding the risk relation.
Driving scenes repeatedly combine a limited set of participant, behaviour and road elements, but risk rules require distinct entity, temporal and relational bindings.
Semantic retrieval can therefore match a risk rule while missing the event and relation rules that establish its scene-specific conclusion.

Our method applies the full DRKG rule set to scene facts and returns rules supporting entailed risks, rather than selecting rules by similarity.
Its 100\% recall and applicability follow by construction against the same SWRL-derived reference, not from independent validation of perception or rule correctness.

\subsection{Planning with Scene-Grounded Risk Evidence}
The low applicability in Table~\ref{tab:rule_retrieval} motivates testing scene-grounded risk evidence against semantic retrieval in planning.
Table~\ref{tab:planning_retrieval} reports this comparison with camera observations, ego context, VLM and diffusion planner fixed.

Scene-grounded risk evidence achieved the highest NPS (64.18) and NC (90.29) in Table~\ref{tab:planning_retrieval}.
KnowVal top-16, the strongest semantic RAG setting by NPS, reached 62.88 and 87.53, respectively.
NC was the only component mean to rise relative to this setting.
Because NC and DA gate NPS multiplicatively in Eq.~\eqref{eq:nps}, the higher NC is consistent with the safety-weighted score improvement.

The NPS gain was not accompanied by closer imitation of the reference trajectory.
Scene-grounded risk evidence primarily changes safety-critical behaviour rather than trajectory imitation fidelity.

Larger retrieval sets did not reliably improve planning despite higher risk-rule recall.
KnowVal NPS rose from 62.01 to 62.88, whereas DriveReg changed little, from 62.53 to 62.49.
Both expanded sets lowered risk-rule applicability (Table~\ref{tab:rule_retrieval}), so broader coverage did not consistently translate into planning gains, because the additional rules are often scene-inapplicable.

\subsection{Ablation of Scene-Grounded Risk Evidence Representation}
To separate scene-bound conclusions from rule semantics, we varied the content of $\mathcal{Z}_s$ while holding activated rules and both models fixed.
The VLM received no risk evidence, inferred events and bound risk relations, activated rule descriptions, or the full risk evidence.
Table~\ref{tab:rule_text_ablation} compares these four representations without changing the activated rules.

\begin{table*}[hbtp]
\caption{Selected planning baselines reported on nuReasoning \cite{huang2026nureasoning} and our method.
}
\label{tab:published_planning}
\centering
\footnotesize
\setlength{\tabcolsep}{6pt}
\begin{tabular}{lccccccc}
\toprule
Method & NC $\uparrow$ & DA $\uparrow$ & EP $\uparrow$ & CF $\uparrow$ & HL $\uparrow$ & NPS $\uparrow$ & ADE $\downarrow$ \\
\midrule
UniAD \cite{hu2023uniad} & 88.87 & 87.62 & 89.62 & 92.62 & 48.80 & 55.65 & 2.054 \\
DiffusionDrive \cite{liao2025diffusiondrive} & 90.22 & 88.25 & 90.46 & 94.96 & 51.96 & 57.86 & 1.930 \\
AutoVLA \cite{zhou2025autovla} & 90.92 & 86.48 & 89.33 & 99.90 & 49.89 & 59.05 & 2.063 \\
SpanVLA \cite{zhou2026spanvla} & 93.78 & 88.35 & 85.72 & 99.80 & 49.13 & 60.59 & 1.890 \\
Alpamayo-1.5 (zero-shot)  \cite{wang2025alpamayo} & 90.26 & 86.13 & 86.51 & 97.93 & 33.79 & 50.45 & 2.925 \\
nuVLA (planning only) \cite{huang2026nureasoning} & 94.87 & 92.10 & 87.38 & 99.70 & 55.22 & 64.98 & 1.937 \\
\midrule
Ours (single-view VLM) & 90.29 & 92.91 & 89.79 & 90.29 & 60.17 & 64.18 & 1.728 \\
\bottomrule
\end{tabular}
\end{table*}

\begin{figure*}[thbp]
    \centering
    \includegraphics[trim=11bp 236bp 24bp 49bp, clip, width=0.98\linewidth]{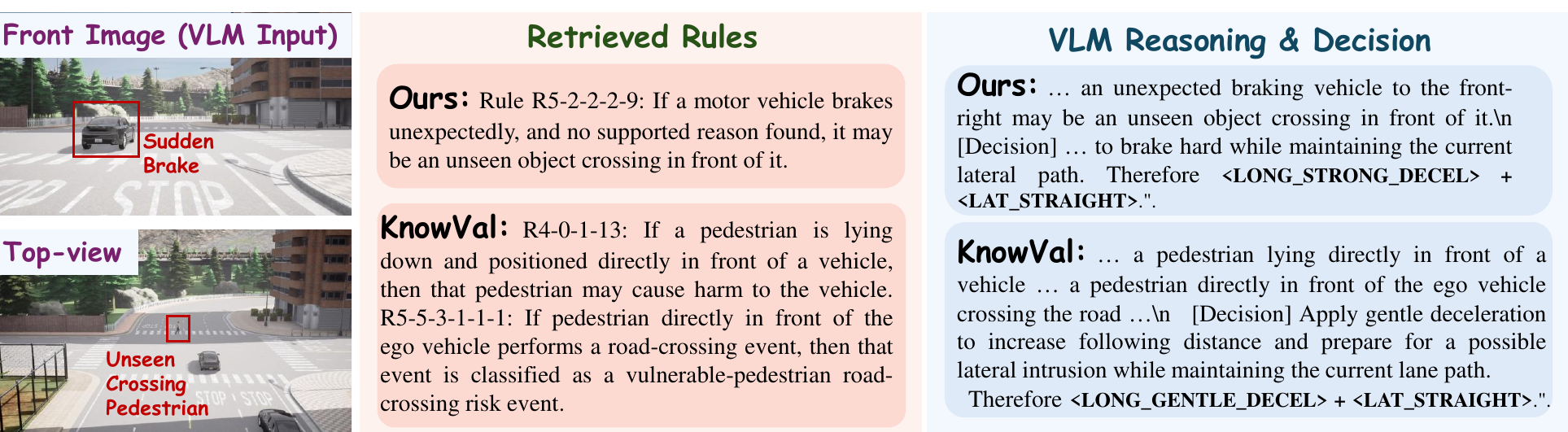}
    \caption{Qualitative comparison in an occluded pedestrian-crossing scene.
    The front camera shows a vehicle braking suddenly, whereas the top view reveals a pedestrian hidden from the ego view.
    Our method identifies the applicable risk rule and selects strong deceleration.
    The KnowVal-style baseline retrieves semantically related pedestrian rules (2 of top-5 are shown) and selects gentle deceleration.}
    \label{fig:example}
    \vspace{-0.3cm}
\end{figure*}

Inferred results alone reached 63.22 NPS, slightly above 62.68 with activated rule descriptions alone (Table~\ref{tab:rule_text_ablation}).
With the activated rules fixed, inferred results supplied recognized events and risk-source and risk-target bindings, whereas descriptions supplied general rule semantics.
This modest difference suggests that explicit scene results add value beyond rule text.

Combining inferred results with rule descriptions yielded the highest NPS (64.18) across the four ablations.
Inferred events and relations supplied scene-specific bindings, while activated rule descriptions explained the risk type and rationale.
Their complementary roles support the evidence representation.

Table~\ref{tab:published_planning} provides context for planner performance rather than a direct leaderboard comparison.
Published rows use the official test set, while ours uses the evaluation split defined in the Dataset subsection.
Our single-view VLM planner reached 64.18 NPS, 92.91 DA and 1.728 ADE, with CF at 90.29, establishing that the downstream planner is competitive enough for controlled risk-information experiments.

\subsection{Qualitative Results}
Fig.~\ref{fig:example} illustrates an occluded pedestrian-crossing risk in a complex driving scene.
The front camera shows another vehicle braking suddenly, while the top view reveals a pedestrian hidden from the ego view.
Risk inference proceeds from braking recognition through visible-cause checking, unexpected-event classification and hidden-cause inference to an ego-risk conclusion.
Our method completed this chain and activated the applicable rule, leading the VLM to select strong longitudinal deceleration while maintaining its lane.
KnowVal instead detected the crosswalk in the image and used it as a keyword to retrieve multiple pedestrian-related rules.
It did not check their scene-specific antecedents, and the resulting gentle deceleration left the immediate occluded-crossing risk insufficiently addressed.

\section{CONCLUSIONS}

Semantic relevance alone cannot establish a scene-specific driving risk because the rule's antecedents must be supported by current scene facts.
Our framework makes this applicability check explicit through DRKG fact instantiation and SWRL entailment, supplying recognized events, directed risk relations and activated rule descriptions to the VLM and diffusion planner.
On nuReasoning, broader semantic retrieval increased risk-rule recall while reducing scene applicability.
Scene-grounded evidence improved NC from 87.53 to 90.29 and NPS from 62.88 to 64.18 relative to the strongest semantic-retrieval condition.
The evidence ablation further showed that scene-bound conclusions and rule semantics contribute complementary information to planning.

The framework relies on structured perception for scene-fact instantiation, and noisy or missed observations can change which SWRL antecedents are satisfied.
It can entail only risk forms covered by the DRKG rules, so incomplete rule coverage limits the hazards represented in its evidence.
Future work will improve robustness to noisy perception, extend the DRKG with broader, automatically acquired risk knowledge, and enhance the generalization by VLMs.






\bibliographystyle{IEEEtran}
\bibliography{ref}

\section*{Supplementary Material}
\subsection*{DRKG Construction and Coverage}
We constructed the DRKG from a hierarchical catalogue of driving-risk scenarios.
Scenarios with similar triggering conditions, interacting entities and event sequences were grouped into risk types.
We mapped the required scene entities and relations to the five ontology modules, then encoded conditional event and risk derivations as SWRL rules.
Six experts reviewed the scenario grouping, ontology and rules for semantic consistency.

Of the 177 SWRL rules in the DRKG, 39 are risk-specific and cover 27 risk types.
Table~\ref{tab:drkg_coverage} reports the number of risk types and rules in each category.
Vars. and Atoms give the mean numbers of unique variables and semantic antecedent atoms per rule, with ranges in parentheses.
Atom counts exclude assertions used only for scene or temporal membership.

\begin{table}[H]
\caption{Coverage and complexity of risk-specific SWRL rules.}
\label{tab:drkg_coverage}
\centering
\footnotesize
\setlength{\tabcolsep}{2.5pt}
\begin{tabular}{@{}>{\raggedright\arraybackslash}p{2.35cm}cccc@{}}
\toprule
Risk category & Types & Rules & Vars. & Atoms \\
\midrule
Vehicle interaction & 8 & 13 & 6.38 (4--11) & 11.08 (6--18) \\
Vulnerable road users & 4 & 6 & 4.33 (4--6) & 7.67 (7--10) \\
Road geometry and traffic conditions & 11 & 14 & 6.00 (5--9) & 10.64 (7--20) \\
Occlusion and blind-spot risk & 4 & 6 & 4.67 (4--6) & 9.00 (7--11) \\
\midrule
Total & 27 & 39 & 5.67 (4--11) & 10.08 (6--20) \\
\bottomrule
\end{tabular}
\end{table}

\end{document}